\documentclass[conference]{IEEEtran}
\usepackage{times}
\usepackage[letterpaper, left=0.75in, right=0.75in, bottom=0.75in, top=1in]{geometry}

\usepackage[numbers]{natbib}
\usepackage{multicol}
\usepackage{graphicx}
\usepackage[normalem]{ulem}
\usepackage{hyperref}
\usepackage{amsmath}
\usepackage{amssymb}
\usepackage{gensymb}
\usepackage{algorithm}
\usepackage[noend]{algpseudocode}
\usepackage{caption}

\usepackage{subcaption}
\usepackage{color}
\usepackage{xcolor}

\begin{document}

% paper title
% \title{Bootstrapping Motor policies with Motion Planning}
%\title{Bootstrapping And Optimizing Motor Policies with Uncertain Kinematic Models \gdknote{Shorter: Bootstrapping Motor Skill Learning with Motion Planning}}
\title{Bootstrapping Motor Skill Learning\\ with Motion Planning}
% You will get a Paper-ID when submitting a pdf file to the conference system
\author{Ben Abbatematteo$^*$, Eric Rosen$^*$, Stefanie Tellex, George Konidaris  \\
  Department of Computer Science\\
  Brown University\\
  Providence, RI \\ 
  \texttt{ \{babbatem,er35,stefie10,gdk\}@cs.brown.edu} \\}
\thanks{These two authors contributed equally}

%\author{\authorblockN{Michael Shell}
%\authorblockA{School of Electrical and\\Computer Engineering\\
%Georgia Institute of Technology\\
%Atlanta, Georgia 30332--0250\\
%Email: mshell@ece.gatech.edu}
%\and
%\authorblockN{Homer Simpson}
%\authorblockA{Twentieth Century Fox\\
%Springfield, USA\\
%Email: homer@thesimpsons.com}
%\and
%\authorblockN{James Kirk\\ and Montgomery Scott}
%\authorblockA{Starfleet Academy\\
%San Francisco, California 96678-2391\\
%Telephone: (800) 555--1212\\
%Fax: (888) 555--1212}}

% avoiding spaces at the end of the author lines is not a problem with
% conference papers because we don't use \thanks or \IEEEmembership

% for over three affiliations, or if they all won't fit within the width
% of the page, use this alternative format:
% 
%\author{\authorblockN{Michael Shell\authorrefmark{1},
%Homer Simpson\authorrefmark{2},
%James Kirk\authorrefmark{3}, 
%Montgomery Scott\authorrefmark{3} and
%Eldon Tyrell\authorrefmark{4}}
%\authorblockA{\authorrefmark{1}School of Electrical and Computer Engineering\\
%Georgia Institute of Technology,
%Atlanta, Georgia 30332--0250\\ Email: mshell@ece.gatech.edu}
%\authorblockA{\authorrefmark{2}Twentieth Century Fox, Springfield, USA\\
%Email: homer@thesimpsons.com}
%\authorblockA{\authorrefmark{3}Starfleet Academy, San Francisco, California 96678-2391\\
%Telephone: (800) 555--1212, Fax: (888) 555--1212}
%\authorblockA{\authorrefmark{4}Tyrell Inc., 123 Replicant Street, Los Angeles, California 90210--4321}}

\maketitle
\def\thefootnote{*}\footnotetext{These authors contributed equally to this work}
\begin{abstract} 
Learning a robot motor skill from scratch is impractically slow; so much so that in practice, learning must be bootstrapped using  a good skill policy  obtained from human demonstration. However, relying on  human demonstration necessarily degrades the autonomy of robots that must  learn a wide variety of skills over their operational lifetimes. We propose using kinematic motion planning as a completely autonomous, sample efficient way to bootstrap motor skill learning for object manipulation. We demonstrate the use of  motion planners to bootstrap  motor skills in two complex object manipulation scenarios with different policy representations: opening a drawer with a dynamic movement primitive representation, and closing a microwave door with a deep neural network policy. We also show how our method can bootstrap a motor skill for the challenging dynamic task of learning to hit a ball off a tee, where a kinematic plan based on treating the scene as static is insufficient to solve the task, but sufficient to bootstrap a more dynamic policy. In all three cases, our method is competitive with human-demonstrated initialization, and significantly outperforms starting with a random policy. This approach enables robots to to efficiently and autonomously learn motor policies for dynamic tasks without human demonstration.

\end{abstract}

\IEEEpeerreviewmaketitle

\section{Introduction}
% In order for manipulator robots to be useful, they require \banote{skills for interacting} \sout{motor policies that let them interact} with the environment. For example, a robot butler may need to open a drawer to get utensils for a table. Also, if the drawer becomes more difficult to open due to rust, the robot should be able to adapt its motor behavior based on the environment. How can robots do this in an efficient manner?

%\banote{TODO: quantitative real t-ball} \\
%\banote{TODO: motion planning disclaimer} \\
%\banote{TODO: model noise ablation?} \\

\begin{figure}
\centering
\begin{subfigure}[b]{0.9\linewidth}
\centering
   \includegraphics[width=0.8\linewidth]{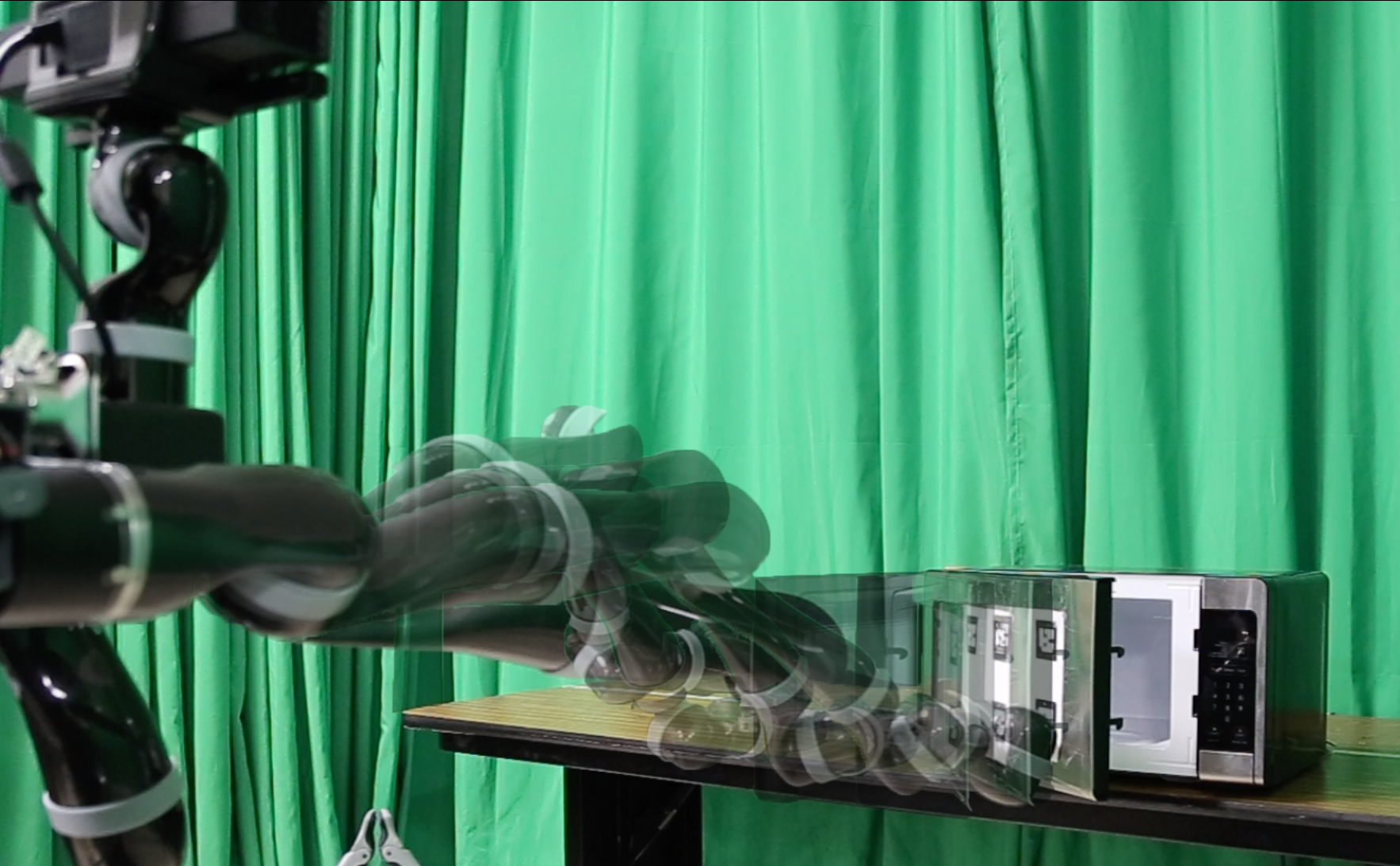}
   \caption{}
   \label{fig:Ng1} 
\end{subfigure}

\begin{subfigure}[b]{0.9\linewidth}
\centering
   \includegraphics[width=0.8\linewidth]{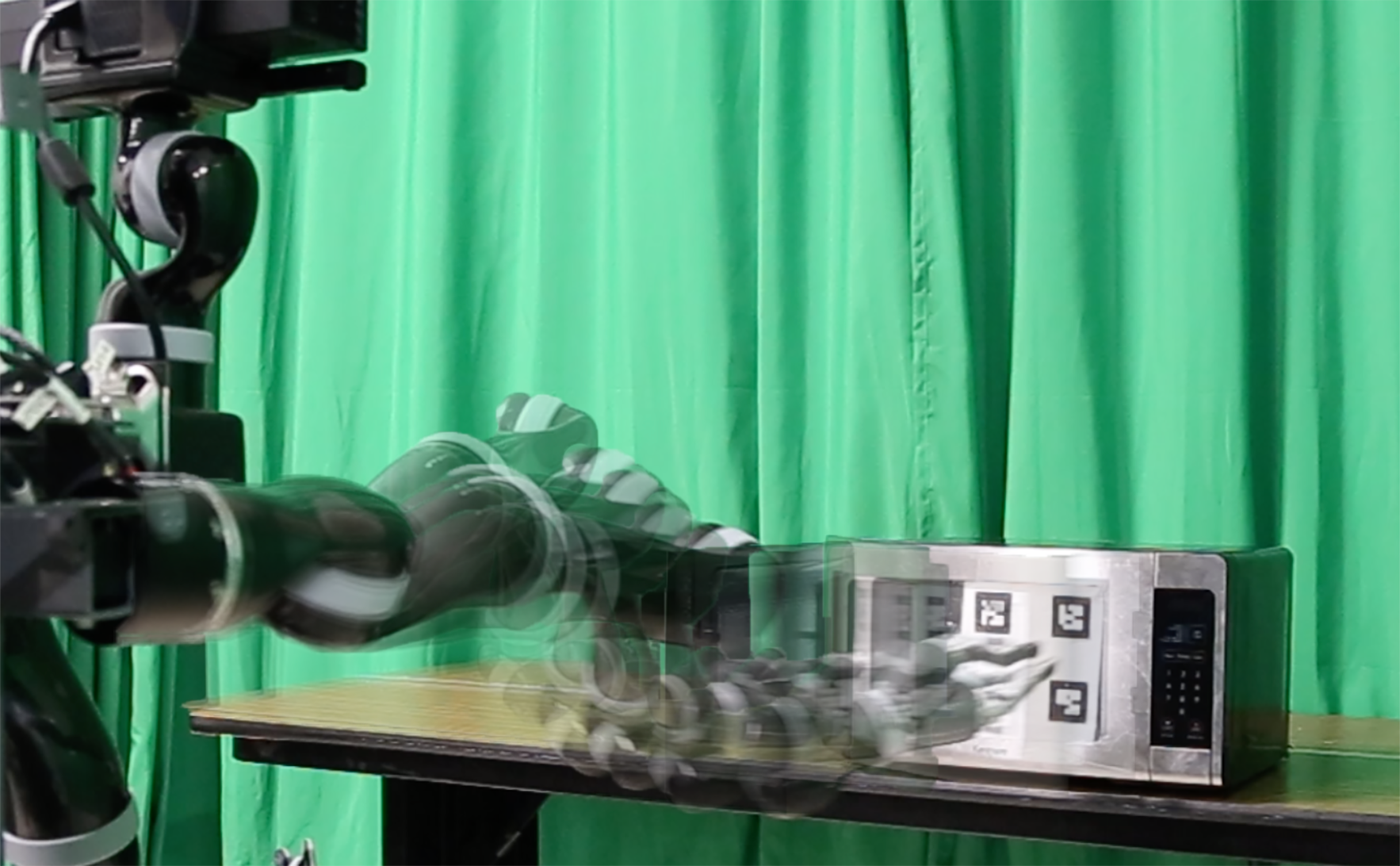}
   \caption{}
   \label{fig:Ng2}
\end{subfigure}

\caption{A robot using our method to autonomously learn to close a  microwave.  (a) The robot uses a motion planner to generate an initial attempt at closing the microwave door using a kinematic model of the microwave. The resulting plan is unable to fully close the microwave door because of the robot's limited reach. (b) After bootstrapping a motor skill with the trajectory from (a), the robot learns a motor skill that gives the door a push, exploiting its dynamics to fully close the microwave.  %\stnote{I think this would look better if you recropped the images so that the Kinect wasn't in the scene (e.g., take a chunk off the top), and then made the image width equal to line width for both.  There will be less green and the microwave itself will be bigger.   I also wonder if it would be clearer to show the first one at just the end state, to show it doesn't succeed at the task... }
}
\label{demopic}
\end{figure}
 Robots require motor policies for interacting with objects in their environment. For example, a robot butler may need a motor skill that enables it to open a drawer to fetch utensils for a table, for setting each element of the table, and for pouring wine.
    While it is safe to assume that a robot will have an accurate kinematic model of its own body, it is unlikely to have a dynamics model of every  object it will ever encounter. This lack of knowledge means that the robot will have to learn how to manipulate the world around it \cite{kroemer2019review}.
    
    Reinforcement learning (RL) provides a framework for robots to acquire motor policies without explicitly modeling the unknown world, but model-free RL methods like policy search \cite{deisenroth2013survey} have high sample-complexity, and often fail to learn a reasonable policy from random initialization. Supervised approaches for policy learning like Learning From Demonstration (LfD) \cite{argall2009survey} can encode human prior knowledge by imitating expert examples, but do not support optimization in new environments. Combining RL with LfD is a powerful method for reducing the sample complexity of policy search, and is often used in practice \cite{levine2013guided, rajeswaran2017learning,Zhu-RSS-18,2018arXiv180510413C}. However, this approach typically requires a human demonstrator for initialization, which fundamentally limits the  autonomy, and therefore utility, of a robot which may need to acquire a wide range of motor skills over its operational lifetime. More recently, model-based control techniques (including Model Predictive Control \cite{kahn2017plato, pan2017agile} and LQR \cite{levine2013guided}) have been proposed as exploration methods for policy search; these methods still require human demonstrations or complete dynamic models of both the robot and every object in the scene.   %\stnote{Can you push the reviews into the repo so we have them?  Does it make sense to have another sentence here about the papers reviewers mentioend that had other kinds of automatic demonstrations?} \ernote{Addressed below}
    %with which to initialize learned models in the absence of known scene dynamics. 
    
    We propose the use of kinematic motion planning to initialize motor skill policies. While previous work have leveraged sample-based motion planners for learning motor skills \cite{tosun2019pixels,jurgenson2019harnessing,jiang2019task}, they only focus on either free-space motions or do not learn a closed-loop controller. To our knowledge, this is the first use of motion planning to provide initial demonstrations for learning closed-loop motor skill policies by leveraging estimated object kinematics.
    Motion planning algorithms 
    %a solution that has the robot use knowledge of its own body (kinematic model) to bootstrap motor policies with motion planning. Motion planning algorithms such as RRT*, CHOMP, and STOMP 
    generate collision-free behavior and generalize to novel scenarios when the robot has a good kinematic model of itself and the object it aims to manipulate, making it useful for tasks like pick-and-place. We show that given a (potentially approximate, and readily estimated) kinematic description of the environment and the robot, off-the-shelf motion planning algorithms can generate feasible (potentially successful but inefficient) initial trajectories (Figure \ref{fig:Ng1}) to bootstrap an object-manipulation policy that can subsequently be optimized using policy search (Figure \ref{fig:Ng2}). This framework enables the robot to automatically produce its own demonstrations for effectively learning and refining object manipulation policies. Our work enables the robot to realize the benefits of an initial demonstration fully automatically using kinematic planning, requiring no human involvement.
    
    % To evaluate our method, we compared our method of bootstrapping motor policies with a motion planner against human demonstrations in both simulated and real robots for three different manipulation tasks: closing a microwave, opening a drawer, and hitting a ball off a tee. Across all three tasks, we found that while bootstrapping with the human demonstration provided an initially better policy than bootstrapping with a motion planner, both policies converged to similarly optimal policies after reinforcement learning was used to refine the policies. Our findings give credence to using motion planners to bootstrap motor policies, and suggest a path forward towards autonomously acquiring motor policies without a human-in-the-loop. 
    
    To evaluate our method, we used two different motor policy classes (Dynamic Movement Primitives (DMPs) \cite{ijspeert2003learning} and deep neural networks \cite{JMLR:v17:15-522}). We chose these two different motor policy classes because deep neural networks are extremely expressive in what policies they can represent, but are extremely sample inefficient compared to structured motor primitives like DMPs, and we aim to evaluate how our method performs in both contexts. Using these motor policy classes, we compared bootstrapping with motion planning against learning from scratch in three simulated experiments, and against human demonstrations in real hardware experiments. Human demonstrations provide a baseline for how effective these motor policies can do when bootstrapped with high-quality demonstrations, and learning from scratch provides a baseline for how difficult the task is without any prior information about the task.% \stnote{This sort of feels like a list of experiments without a lot of context, can you rephrase to introduce why each experiment is important or what it is telling us about themodel?}
    In the first two experiments---opening a drawer with a dynamic movement primitive representation, and closing a microwave door with a deep neural network policy---we show that motion planning using a kinematic model produces a reasonable initial policy,although suboptimal compared to a supervised human demonstration, that learning adapts to generate efficient, dynamic policies that exploit the dynamics of the object being manipulated.
    We also show how our method can bootstrap a motor skill for the challenging dynamic task of learning to hit a ball off a tee, which involves precise and agile movement. In that case, treating the objects in the scene as static and applying kinematic motion planning succeeds in generating a policy that makes contact with the ball, which is sufficient to bootstrap a more dynamic policy that learns to hit the ball several feet. Our method is competitive with human-demonstrated initialization, but requires no human demonstration.  It serves as a suitable starting point for learning, and significantly outperforms starting with a random policy. This approach enables robots to efficiently and autonomously learn motor policies for dynamic tasks without human demonstration.
    
    In summary, our contributions are: 
    % a)  a fully-autonomous paradigm for policy search, in which an agent first uses goal-directed kinematic planning to devise feasible static solution trajectories for itself and objects in a scene. b) a novel algorithm that autonomously generates a set of initial demonstrations via kinematic planning in object and robot configuration spaces. c) An empirical evaluation of this algorithm in which we employ model-free policy optimization after bootstrapping, demonstrating that our method's performance is comparable to human expert demonstration and superior to random initialization in hardware and simulation tasks: closing a microwave, opening a drawer, and hitting a ball off a tee. 
    \begin{enumerate}
        \setlength{\itemsep}{0pt}%
        \setlength{\parskip}{0pt}%
        \item A fully-autonomous paradigm for policy search, in which an agent first uses goal-directed kinematic planning to devise feasible solution trajectories for itself and objects in a scene.
        \item  A novel algorithm that autonomously generates a set of initial demonstrations for object manipulation via kinematic planning in object and robot configuration spaces.  
        \item An empirical evaluation of this algorithm in which we employ model-free policy optimization after bootstrapping, demonstrating that our method's performance is comparable to human expert demonstration and superior to random initialization in hardware and simulation tasks: closing a microwave, opening a drawer, and hitting a ball off a tee. 
    \end{enumerate}

\section{BACKGROUND}
%\stnote{I think an intro paragraph is really important here, setting up the tree of problems and solutions.} \\
%\banote{how's this} \\
%\banote{also, how is order of subsections? } \\
%\ernote{I like the ordering!}
%\gdknote{perfecto}

Our goal is to efficiently and autonomously learn robot motor skill policies. To do so, we develop an approach that uses kinematic motion planning to generate initial trajectories, fits a policy to those trajectories using behavioral cloning, and subsequently optimizes that policy via policy search. We now briefly describe policy search, policy representations, learning from demonstration, and motion planning. 

\subsection{Policy Search}
%Several examples of initializing policies with human demonstrations exist in the literature \banote{cite DMP things, RMP things, Imitation learning w/ NN things?}.
Policy search methods \cite{deisenroth2013survey} are a family of model-free reinforcement learning algorithms that search within a parametric class of policies to maximize reward. Formally, given a Markov Decision Process $\mathcal{M} = \langle \mathcal{S},\mathcal{A},\mathcal{R},\mathcal{T},\gamma \rangle$, the objective of policy search is to maximize the expected return of the policy $\pi_\theta$:
        \begin{equation}
        \max_\theta \mathop{\large{\mathbb{E}}}_{\mathcal{M},\pi_\theta} \left[ \sum_{t=0}^{T} \gamma^t r_t \right]. \label{eq: rl}
        \end{equation}
        These approaches can learn motor skills through interaction, and therefore do not require an explicit environment model, and are typically agnostic to the choice of policy class (though their success often depends on the policy class having the right balance of expressiveness and compactness). However, their model-free nature leads to high sample complexity, which often makes them infeasible to apply directly to robot learning problems.
        
        \textbf{Policy representation} describes the class of functions used as the mapping from states to actions. 
        Our approach is agnostic to policy representation; we demonstrate our approach enables efficient learning using two different common policy representations: Dynamic Movement Primitives and neural network controllers. 
        
        Dynamic Movement Primitives \cite{ijspeert2003learning, peters2008natural} are a description of a non-linear second-order differential equation that exhibits attractor dynamics modulated by a learnable forcing function. 
        DMPs are a popular representation for motor policies because they are parameter-efficient, can express both point and limit cycle attractors, enable real-time computation, and exhibit temporal invariance that does not effect the attractor landscape.
        %\ernote{ More formally, a DMP is defined as the following expression:
        %$$\tau \ddot{y}=\alpha_{y}(\beta_{y}(g-y)-\dot{y}) + f$$
        %Where $y$ is the system state, $g$ is the goal system state, $\alpha$ and $\beta$ are gain parameters, $\tau$ is a temporal scaling parameter, and $f$ is a nonlinear forcing function. Ignoring $f$, the above equation describes a PD control signal, which draws the system state to $g$ with simple spring dampening properties. The forcing function $f$ is typically represented by a linear combination of activation functions, making it easy to initialize from a demonstration.} 
        We refer the reader to the work of \citet{ijspeert2003learning} for a more  formal introduction to DMPs. 
        If we have $n$ joints we wish to control, we can model control for each joint independently with $n$ DMPs for each one. 
        Therefore, the multiple joints of a robot are only coupled through time, which makes this representation very compact.   
        %\gdknote{Might be helpful to describe the ``shape'' part explicitly, talk about how its easy to initialize because its linear, and then about how it decouples the joints from each other, and only couples them via the time, which makes this representation very compact.}
        
        % In order to shape the DMP to produce trajectories similar to an initial demonstration, we can use locally weighted regression to learn the forcing function. More details on using locally weight regression for fitting DMPs can be found in \cite{schaal2002scalable}.
        
        Neural network controllers have received significant attention in recent years; they are able to learn hierarchical feature representations for approximating functions (in our case, motor skills) operating on high-dimensional input such as robot sensor data. They are more expressive than restricted policy classes such as DMPs and can operate directly on high-dimensional state spaces (e.g. images), yet they typically exhibit higher sample complexity \cite{JMLR:v17:15-522}. 
    
    % \subsection{Learning from Demonstration}
    \textbf{Learning from Demonstration} methods
     \cite{argall2009survey, ravichandar2019recent} broadly consist of two families of approaches that either mimic (Behavioral Cloning) or generalize (Inverse Reinforcement Learning) the exemplified behavior. Inverse reinforcement learning methods seek to estimate a latent reward signal from a set of demonstrations; we assume a given reward function, and omit a discussion of inverse reinforcement learning methods here. 
        
        Behavioral cloning methods
        \cite{atkeson1997robot, pastor2009learning, ho2016generative} attempt to directly learn a policy that reproduces the demonstrated policies.
        Given a dataset of expert demonstrations $D$, the objective of behavioral cloning is:
        $\max_\theta \sum_{(s,a) \in D} \pi_\theta(a | s).$

        These methods often result in policies with undesirable behavior in states not observed during demonstrations, though this can  addressed with interactive learning \cite{ross2011reduction, nair2017combining, sun2017deeply}. In our approach, the existence of a reward function enables the agent to learn robust behavior in states outside of the initial training distribution. Moreover, our experiments demonstrate our approach's ability to extrapolate beyond suboptimal initial demonstrations. 
        
        Many approaches investigate the incorporation of human-provided demonstrations into policy search to  drastically reduce sample complexity via a reasonable initial policy and/or the integration of demonstrations in the learning objective \cite{peters2008natural, kang2018policy, 2018arXiv180510413C, rajeswaran2017learning, Zhu-RSS-18, vecerik2017leveraging, levine2013guided}. 
        %and we are agnostic to the choice of policy optimization algorithm.  \gdknote{the last sentence fragment? does not follow at all from the first bit?}
    
    \subsection{Motion Planning}
        
        The pose of an articulated rigid body can defined by the state of each of its movable joints. The space of these poses is called the configuration space $\mathcal{C}$ \cite{lozano1981automatic}. Motion planning is the problem of finding a path (sequence of poses) through configuration space such that the articulated object is moved to a desired goal configuration, without encountering a collision.
        
        While there exist many different families of motion planning algorithms, such as geometric, grid-based, and probabilistic road maps \cite{lavalle2006planning}, they all operate in a similar fashion: given a configuration space $\mathcal{C}$ and start and goal joint configurations $q_{0}, q^* \in \mathcal{C}$, return a valid path of joint configurations $\{q_{t}\}_{t=0}^{T}$ between the start and end configurations. We focus on sample-based motion planning approaches.
        
        Probabilistic motion planners provide a principled approach for quickly generating collision-free robot trajectories.
        % by removing joint configurations from the search space that would be in collision with the environment. They are also able to quickly generate trajectories by using heuristics to guide samples towards joint configurations that are nearby the goal. 
        However, online replanning is expensive, and kinematic motion planners are only as effective as their kinematic models are accurate: they generate trajectories directly, and thus cannot be improved through subsequent interaction and learning. Furthermore, kinematic planners produce trajectories that only account for kinematics, not dynamics: they explicitly do not account for forces involved in motion, such as friction, inertial forces, motor torques, etc, which are important for effectively performing contact-rich, dexterous manipulation. 
        
        The process of computing the position and orientation $p \in SE(3)$ of a link in a kinematic chain for a given joint variable setting (a point in configuration space) is termed \textit{forward kinematics}. Inversely, computing a configuration to attain a specific end effector pose $p$ is termed \textit{inverse kinematics}. We denote the forward kinematics functions $p=f(q)$. 
        
        % While we can actuate robot joints by using its motors, when manipulating the degrees of freedom of an object, we must use the robot's body to indirectly actuate them. For example, a robot wishing to open a cupboard must first grasp the cupboard's door before it can manipulate its hinge joint. Nonetheless, we can form and exploit kinematic plans in object configuration space for efficient exploration. 

\section{Bootstrapping Skill Learning with Motion Planning}
Our methodology is inspired by how humans generate reasonable first attempts for accomplishing new motor tasks. When a human wants to learn a motor skill, they do not start by flailing their arms around in a random fashion, nor do they require another person to guide their arms through a demonstration. Instead, they  make a rough estimate of how they want an object to move and then try to manipulate it to that goal. For example, before being able to drive stick shift, a human must first learn how to manipulate a gear shifter for their car. Just by looking at the gear shifter, humans can decide (1) what they should grab (the shaft), (2) where they want the shaft to go (positioned in a gear location), and (3) how the shaft should roughly move throughout the action (at the intermediate gear positions). Similarly, a robot that has a good kinematic model of itself, and a reasonable kinematic model of the object it wishes to manipulate, should be able to form a motion plan to achieve the effect it wishes to achieve. 
    
    That plan may be inadequate in several ways: its kinematic model may be inaccurate, so the plan does not work; object dynamics (like the weight of a door, or the friction of a joint) may matter, and these are not represented in a kinematic model; and a feasible and collision-free kinematic trajectory may not actually have the desired effect when executed on a  robot interacting with a real (and possibly novel) object. But such a solution is a \textit{good start}; we therefore propose to use it to bootstrap motor skill learning. 

\begin{figure}
    \centering
    \includegraphics[width=\linewidth]{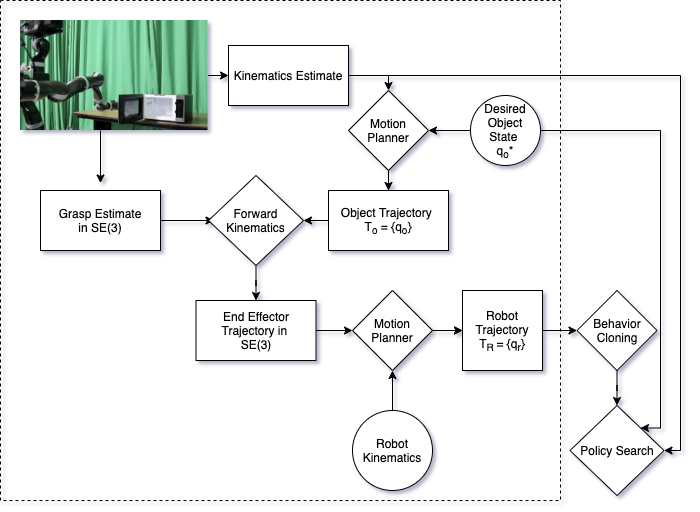}
    \caption{\textbf{System overview} illustrating our proposed framework for generating demonstrations with a motion planner and subsequently performing policy search. The dashed box contains the steps from Algorithm 2.  %\stnote{I think you could cut this if you need space... the pseudocode does it already kinda.}
    }
    \label{fig:overview}
\end{figure}

    Our approach, outlined in Figure \ref{fig:overview}, leverages the (partial) knowledge the robot has about its own body and the object it is manipulating to bootstrap motor skills. Our method first assumes access to the configuration space of the robot, denoted as $\mathcal{C}_{R}$, as well as its inverse kinematics function $f^{-1}_R$. This assumption is aligned with the fact that the robot often has an accurate description of its own links and joints and how they are configured during deployment. However, the world is comprised of objects with degrees of freedom that can only be inferred from sensor data. Therefore, our approach only assumes access to estimated kinematics of the object to be manipulated, in the form of configuration space $\mathcal{C}_{O}$ and forward kinematics $f_O$. Recent work has shown that estimating these quantities for novel objects from sensor data in real environments is feasible \cite{benny, li2019category}, though state-of-the-art estimates still include noise. 
    
    % \ernote{Make a bigger deal how about how  object kinematics are explicitly used for the purpose of a) generating initial trajectories and b) tracking state of object at run time, and that both joints and links are considered.}
    
    Finally, our approach assumes that the task goal can be defined in terms of kinematic states of the robot and environment. Examples of such tasks include pick-and-place, articulated object manipulation, and many instances of tool use. (Note that this requirement fails to capture reward functions  defined in terms of force, for example exerting a specific amount of force in a target location.) Such a goal, together with object and robot kinematic descriptions, enables us to autonomously generate useful initial trajectories for policy search.
    
    Our approach is outlined in Algorithm \ref{algoloop}, and can broken down into five main steps: 1) collect initial trajectories(s) from a motion planner using estimated object kinematics, 2) fit a policy with these initial trajectories, 3) gather rollouts to sample rewards for the current policy based on the kinematic goal, and 4) update the policy parameters based on the actions and rewards, 5) repeat steps 3-4.

\begin{algorithm}
\caption{Planning for Policy Bootstrapping} 
\begin{algorithmic}[1]
\Procedure{PPB}{$C_{R}, f^{-1}_R, C_{O}, f_{O}, q_{O}^*$} 
    \State $D \leftarrow \varnothing$
    \For{$0$ to $N$}   
        \State $D \leftarrow$ InitialMPDemos($C_{R},f^{-1}_R, C_{O},f_{O}, q_{O}^*$) $\cup$ $D$
    \EndFor
    \State $\theta \leftarrow$ FitPolicy($D_{0},...,D_{N}$) 
    \For{$0$ to $E$}     
        \State $T_{0},..,T_{n} \leftarrow$ Rollout($\pi,\theta,q_{O}^*$) 
        \State $\theta \leftarrow$ UpdatePolicy($T_{1},..,T_{n},\theta$) 
    \EndFor
\EndProcedure
\end{algorithmic}
 \label{algoloop}
\end{algorithm}

\begin{algorithm}
\caption{Initial Motion Plan Demos }
\begin{algorithmic}[1]

\Procedure{InitialMPDemos}{$C_{R}, f^{-1}_R, C_{O},f_{O}, q_{O}^*$}
    \State $T_{O} \leftarrow$ MotionPlanner($C_{O}, q_{O}^*$)
    \State $g \leftarrow$ EstimateGrasp($C_{O}$,$f_{O}$)
    \State $eepath \leftarrow$ GraspPath($T_{O},C_{O},f_{O}, g$)
    \State $T_{R} \leftarrow$ MotionPlanner($C_{R}$,  $eepath$, $f^{-1}_R$)
    \State return $T_{R}$
\EndProcedure
\end{algorithmic}
 \label{mpalgo}
\end{algorithm}

\subsection{Initial Trajectories from Motion Planner} 

To fit our policy class, we choose to sample $N$ demonstrations using a motion planner. Our approach for collecting initial demonstrations with a kinematic motion planner is outlined in Algorithm \ref{mpalgo}, and proceeds as follows. First, we use a motion planner to find a path through the object's configuration space $C_{O}$ that moves the object from its initial state to a goal state $q_{O}^*$ using an off-the-shelf motion planner. This produces a joint trajectory in object configuration space, $T_{O}$, which transforms the object from its current joint configuration to the desired one.
        
        % \stnote{An example would help here.  For example, the robot takes a kinematic model of the microwave and plans a joint configuration that would cause the door to open, assuming no noise and the kinematic model is correct.}
        We then estimate a grasp point on the object to designate the contact point for the robot during manipulation. This can be done by either generating candidate grasps using off-the-shelf grasping algorithms \cite{gualtieri2016high, mahler2017dex} or choosing a part semantically. This produces a local 6D pose, $g$, that represents where the robot should  grasp the object during manipulation.   %\stnote{For example, for the microwave, this algorithm would generate a grasp point on the handle of the door?  Anyway an example will help here. }
        
        We then use the grasp point $g$, object joint trajectory $T_{O}$, and the object's forward kinematics $f_{O}$ to generate the series of 6D Cartesian poses that the grasp point $g$ will go through as the object proceeds through $T_{O}$. This produces a series of 6D Cartesian poses, $eepath$, which the robot end effector must go through, assuming a fixed grasp pose to the object.
        
        Finally, we solve for a path in robot joint space that achieves the end effector path in Cartesian space using off-the-shelf sample-based motion planners, using the robot's inverse kinematics $f^{-1}_R$ and the sequence of end-effector poses $eepath$. 
        
        Note that in our experiments, motion plans were generated offline, rather than recomputed online based on the object's tracked state, but online motion planning is a trivial extension.
        
        % It is worth noting that there are other potential methods of using the robot and (estimated) object kinematics to generate trajectories using motion planning. Specifically, since our method uses a fixed grasp between the robot and the object, our method is similar in effect to planning with a closed kinematic chain \ernote{Cite}. However, planning with closed kinematic chains requires motion planners to work with implicit polynomial constraints, which limits the usage of sample-based approaches. Our approach instead first uses motion planning with the estimated object dynamics to produce a trajectory that constrains motion planning with the robot dynamics. \ernote{Does this all make sense, or was it confusing?}
        % \banote{isn't this just closed-chain kinematic planning with extra steps?} \ernote{Good point, I wrote the above paragraph to justify why we don't do closed-chain kinematic planning, thoughts?} \gdknote{I think we can drop the para above, it's a good technical point but a little aside from the narrative}

\begin{figure*}
% \begin{subfigure}[b]{0.32\linewidth}
%          \centering
%          \includegraphics[width=\linewidth]{figures/2-29/deep-microwave.png}
%          \caption{Microwave closing (MLP) }
%          \label{fig:deep-microwave}
%      \end{subfigure}
%  \begin{subfigure}[b]{0.32\linewidth}
%          \centering
%          \includegraphics[width=\linewidth]{figures/2-29/dmp-drawer.png}
%          \caption{Drawer opening (DMP)}
%          \label{fig:dmp-drawer}
%      \end{subfigure}
%      \begin{subfigure}[b]{0.32\linewidth}
%          \centering
%          \includegraphics[width=\linewidth]{figures/2-29/dmp-dynamic.png}
%          \caption{T-ball (DMP) }
%          \label{fig:dmp-dynamic}
%      \end{subfigure}
\begin{subfigure}[b]{0.32\linewidth}
         \centering
         \includegraphics[width=\linewidth]{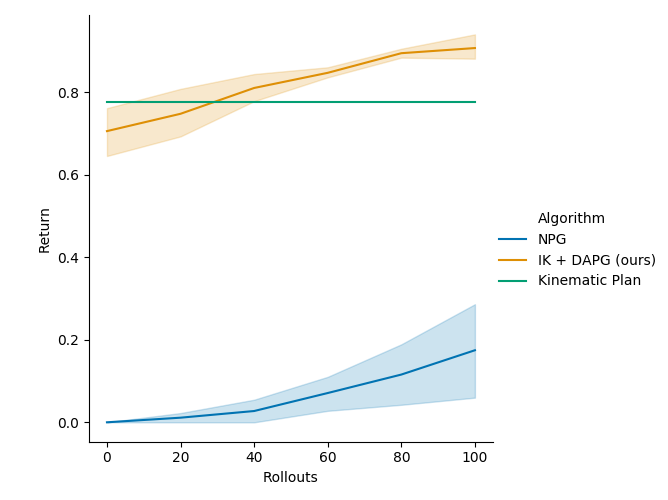}
         \caption{Microwave closing (MLP) }
         \label{fig:deep-microwave}
     \end{subfigure}
 \begin{subfigure}[b]{0.32\linewidth}
         \centering
         \includegraphics[width=\linewidth]{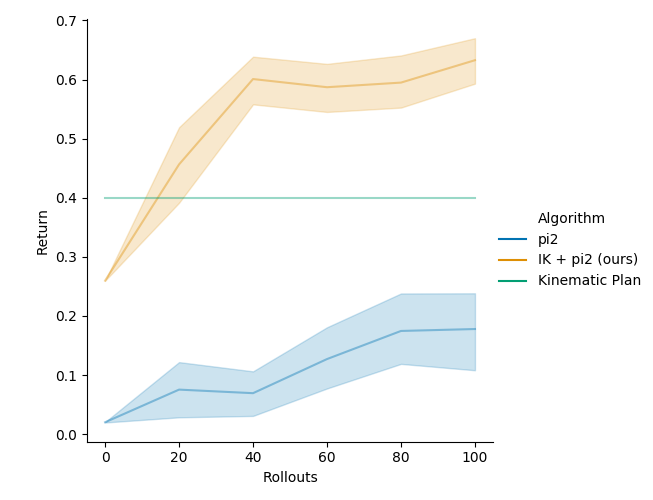}
         \caption{Drawer opening (DMP)}
         \label{fig:dmp-drawer}
     \end{subfigure}
     \begin{subfigure}[b]{0.32\linewidth}
         \centering
         \includegraphics[width=\linewidth]{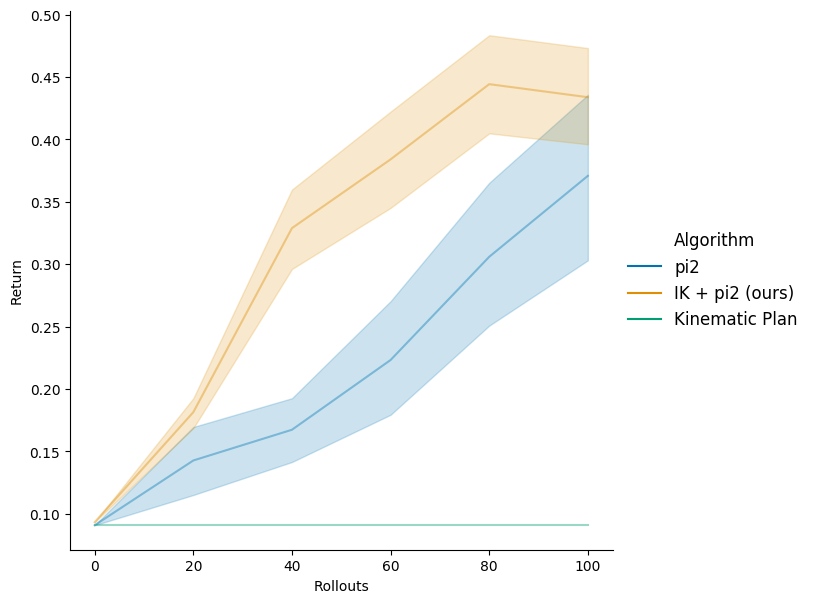}
         \caption{T-ball (DMP) }
         \label{fig:dmp-dynamic}
     \end{subfigure}
        \label{fig:threegraphs}
   \caption{\textbf{Simulation Results.} a) Comparison of our method optimized with DAPG against Natural Policy Gradient starting with a random policy in a microwave closing task using Gaussian multi-layer perception policies. b) Comparison of our method against PI$^2$-CMA starting with a random policy in a drawer opening task with DMP policies. c) Our method compared with PI$^2$-CMA with a initially random policy in t-ball with DMP policies. Results are shown as mean and standard error of the normalized returns aggregated across 20 random seeds.}
   \label{graphy}
\end{figure*}

    % \subsection{Fitting a Policy to a Demonstration}
    \subsection{Fitting a Policy to a Demonstration}
        After collecting initial demonstrations from the motion planner, $D$, we can bootstrap our motor policy by initializing the parameters to the policy $\theta$. We can initialize a parameterized motor policy using any behavioral cloning technique; in practice, for DMPs, we use Locally Weighted Regression \cite{schaal1998constructive}, and for neural networks, we maximize the likelihood of the demonstration actions under the policy. 

    \subsection{Policy Search with Kinematic Rewards}
    \label{sec:ps}
        To improve the motor policies after bootstrapping, we can perform policy search based on the given (kinematic) reward function. Specifically, we choose a number of epochs $E$ to perform policy search for. For each epoch, we perform an iteration of policy search by executing the policy and collecting rewards based on the goal $q_{O}^*$. We define our reward functions using estimated object states $q_{O}$ and goal states $q_{O}^*$, and add a small action penalty. 

\section{EXPERIMENTS}

 The aim of our evaluation was to test the hypothesis that motion planning can be used to initialize policies for learning from demonstration without  human input. We tested this hypothesis in simulation, against learning from scratch, and on real hardware, against human demonstrations, on three tasks: microwave-closing, drawer-opening, and t-ball. We note that we do not show asymptotic performance because our emphasis is on learning on real hardware from a practical number of iterations. All the components of the motion planning problem - state sampler, goal sampler, distance metrics, etc. - are reused between problems without modification.

\subsection{Simulation Experiments}
\label{sec:experiments}
    
     The aim of our evaluation was to test the hypothesis that motion planning can be used to initialize policies for learning from demonstration without  human input. We tested this hypothesis in simulation against learning from scratch, and on real hardware against human demonstrations, on three tasks: microwave-closing, drawer-opening, and t-ball.  
    
    \subsection{Simulation Experiments}
    \label{simexp}
    
         We used PyBullet \cite{coumans2013bullet} to simulate an environment for our object manipulation experiments. 
        We used URDFs to instantiate a simulated 7DoF KUKA LBR iiwa7 arm and the objects to be manipulated, which gave us ground-truth knowledge of the robot and object kinematics.
        For all our simulated experiments, we compared implementations of our method against starting with a random policy.
        
        For all three tasks, the state was represented as $s_t = [q_R, q_{O}]^T$ where $q_R$ denotes robot configuration and $q_{O}$ denotes object configuration. The action space $A$ was commanded joint velocity for each of the 7 motors. 
        The reward at each timestep $r_t$ was given as:
        \begin{equation}
        r_t = -c \ || q_{O}^* - q_{O}||_2^2 - a_t^TRa_t,     \label{eq:r_t}
        \end{equation}
        where $q_{O}$ denotes the object state at time $t$, $q_{O}^*$ denotes desired object state, and $a_t$ denotes the agent's action. We set $c=60$ and R = $I \times 0.001$ for all experiments. As such, maximum reward is achieved when the object is in the desired configuration, and the robot is at rest. 
        % The objective for policy optimization given in Equation \ref{eq: rl}.
        
        Our first simulated task was to close a microwave door, which consisted of three parts: a base, a door, and a handle. 
        % The base and door were connected via a revolute joint, and the door and handle were connected  via a fixed joint. The revolute joint was responsible for determining the angle of the door, where an angle of 90 degrees was completely open and 0 degrees was completely closed. 
        The pose of the handle was used for the EstimateGrasp method in Algorithm 2. The robot was placed within reaching distance of the handle when the microwave door was in an open position, but was too far to reach the handle in its closed configuration. Thus, the agent was forced to push the door with enough velocity to close it. We used Gaussian policies represented as multi-layer perceptrons with two hidden layers of sizes (32,32) in this experiment. The randomly initialized policy was optimized with natural policy gradient \cite{kakade2002natural}. Ten demonstrations were generated by perturbing the start state and initial kinematic plan with Gaussian noise. The behavior cloning was performed by maximizing likelihood over the demonstration dataset for 10 epochs. Our pretrained policy was optimized using Demo Augmented Policy Gradient \cite{rajeswaran2017learning}, which essentially adds the behavior cloning loss to the natural policy gradient loss, annealing it over time. This ensures that the agent remains close to the demonstrations early in learning, but is free to optimize reward exclusively as learning progresses. Results are shown in Figure \ref{fig:deep-microwave}.  

\begin{figure}
    \centering
    \includegraphics[width=\linewidth]{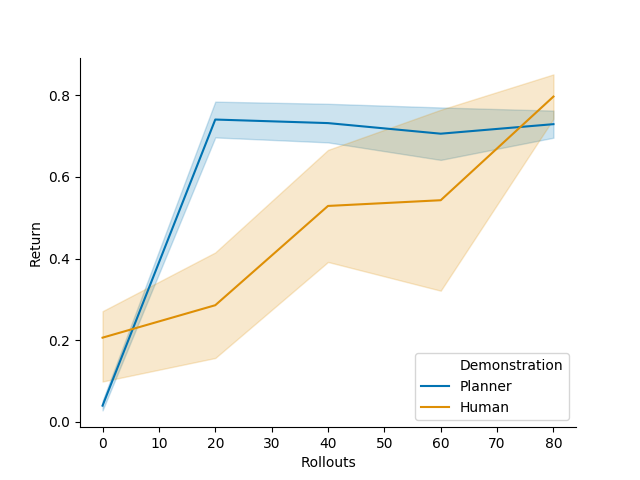}
    \caption{\textbf{Hardware experiment} comparing our initialization scheme with human demonstration. Results are shown as mean and standard error, aggregated across three random seeds. }
    \label{fig:real-microwave}
\end{figure}

The second simulated task was to open a drawer. This task required the agent to grasp the drawer's handle and pull the drawer open.
        % The base and the drawer were connected via a prismatic joint, and the drawer and handle was connected via a fixed joint. The prismatic joint was responsible for determining whether the drawer was open or closed, where an angle of 90 degrees was completely open and 0 degrees was completely closed. 
        Again, the pose of the object's handle was used for EstimateGrasp method in our algorithm. 
        % The drawer handle has a hook shape, requiring the robot to place it's end effector into the handle gap in order to open the drawer from the closed position. 
        In this experiment, we used DMP policies. The weights, goals, and speed parameters of the policies were optimized using PI$^2$-CMA \cite{stulp2012path}. We used 32 basis functions for each of the DMPs. The pretrained policy was initialized using Locally Weighted Regression (LWR) \cite{schaal1998constructive} with a single demonstration. The results of this experiment are shown in Figure \ref{fig:dmp-drawer}. 
        
        The third simulated task was to hit a ball off a tee. The ball started at rest on top of the tee. The pose of the ball was used in the EstimateGrasp method. The object state was defined as the object's $y$ position relative to its initial pose. This experiment again used DMP policies initialized with LWR and optimized with PI$^2$-CMA. The results of this experiment are visualized in Figure \ref{fig:dmp-dynamic}.
        
        The results of our simulated tasks can be found in Figure \ref{graphy}. Across all three tasks, we observe that policies initialized with our method dramatically outperform starting learning with a random policy. This confirms our hypothesis that using motion planning to generate demonstrations significantly speeds the acquisition of motor skills in challenging tasks like articulated object manipulation and t-ball. 
        
        \begin{figure}
     \begin{subfigure}[b]{0.48\linewidth}
         \centering
         \includegraphics[width=\linewidth]{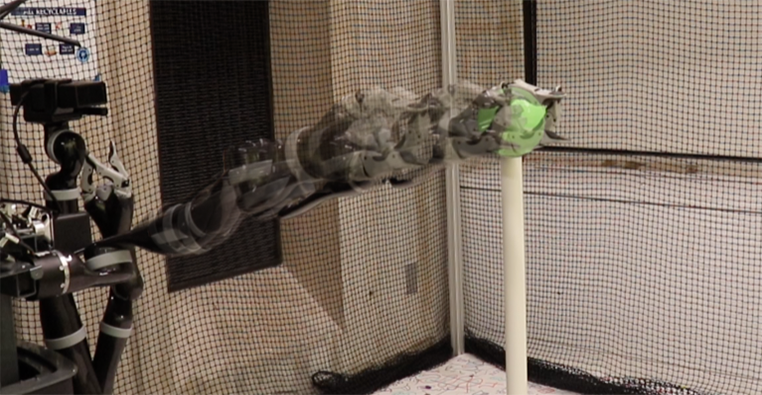}
         \caption{Motion Plan Demonstration}
         \label{fig:mp-demo}
     \end{subfigure}
     \begin{subfigure}[b]{0.48\linewidth}
         \centering
         \includegraphics[width=\linewidth]{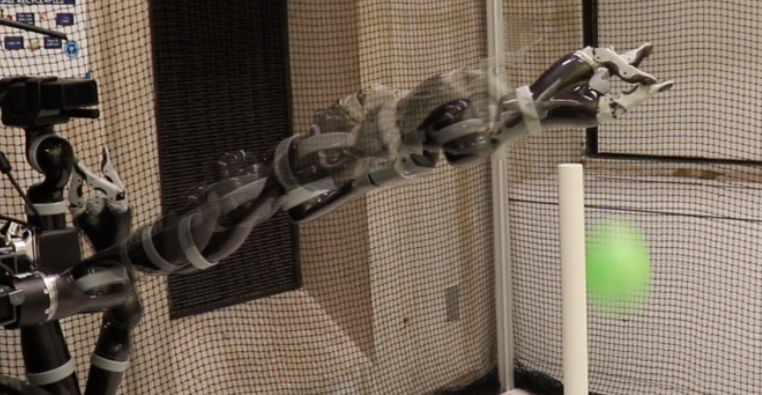}
         \caption{Bootstrapped from (a)}
         \label{fig:mp-learned}
     \end{subfigure}
        \label{fig:ballpic}
   \caption{\textbf{Real-world Ball Hitting} Images comparing an autonomous motion plan demonstration generated from our method vs. a bootstrapped motor skill initialized with that demonstration for a real-world robot hitting a ball off a tee as far as possible. Qualitatively, the bootstrapped motor skill outperforms the initial demonstration by learning to take advantage of the latent tasks dynamics. Videos can be found in our supplemental video. (a) A demonstration provided by the motion planner, which moves linearly towards the ball (b) A motor skill bootstrapped by the motion planner demonstration that learns a agile swooping motion to take advantage of the balls dynamics.}
   \label{fig:real-tball}
\end{figure}
        
        %choose three different tasks: closing a microwave door, closing a drawer, and hitting a ball off a tee. We performed these experiments both on a virtual robot arm, and a real robot arm\banote{revisit}. The virtual environment was done in Pybullet with a Kuka arm \ernote{cite}, and for the real robot experiment, we use a Kinova Movo with a 7DoF Jaco arm. 
        
        %In both the simulated and real scenarios, we use noisy estimates of the object's kineamtic description and accurate models of the robot's kinematics. We represent the kinematic graph susing a URDF \ernote{cite}, which we include as part of the supplementary material \ernote{do this} \banote{remove noisy estimate things?}. Therefore, the robot and target objects are considered instantiations of their respective URDFs, which have specified joint state values\banote{this is weird?}. \sout{It is this representation that we use for our kinematic motion planning and defining the reward signal.}
        
        %For the simulated tasks, we had ground-truth access to the object's kinematic description and joint states, which allowed us to collect a noiseless signal of the object state for reinforcement learning. We also conducted experiments where we added increasing amounts of noise to the estimated kinematic state \banote{not yet we didn't}. 

        \subsection{Real-world Experiments}
        For all our real-world experiments, we used a 7DoF Jaco arm \cite{campeau2019kinova} to manipulate objects (Figure \ref{demopic}). 
        % For the purpose of motion planning with the real-world objects, we represented them with URDFs. 
        We used ROS and MoveIt!\cite{chitta2012moveit} as the interface between the motion planner (RRT* \cite{karaman2011anytime} in our experiments) and robot hardware. For all  real-world experiments, we compared implementations of our method against bootstrapping with a human demonstration, which we supplied. We acknowledge this potential bias in expert trajectories, and qualify our decision by only training on human demonstrations that at least accomplished the task. To collect human demonstrations, we had an expert human teleoperate the robot with joystick control to perform the task. For all tasks, the state space, action space, and reward were defined in the same way as in our simulated results (Section \ref{simexp}). Both experiments used DMP policies initialized with LWR \cite{schaal1998constructive} and optimized with PI$^2$-CMA \cite{stulp2012path} with 10 basis functions for each of the DMPs.
        %In both experiments, we use DMPs as our policy class, where the weights, goals and speed parameters of the policies were optimized using PI$^2$-CMA. \cite{stulp2012path}. We used 10 basis functions for each of the DMPs. The pretrained policy was initialized using LWR \cite{schaal1998constructive} with a single demonstration

        Our first real-world task was to close a microwave door, similar to the one described in our simulated domain (Section \ref{simexp}). As in the simulated microwave task, we used the pose of the handle for the EstimateGrasp method in Algorithm \ref{mpalgo}, and also the robot was similarly placed such that it was forced to push the door with enough velocity to close. %We use DMPs as our policy class, where the weights, goals and speed parameters of the policies were optimized using PI$^2$-CMA. \cite{stulp2012path}. We used 10 basis functions for each of the DMPs. The pretrained policy was initialized using LWR \cite{schaal1998constructive} with a single demonstration.
        We placed an AR tag on the front-face of the microwave to track the microwave's state using a Kinect2. 
        % We then continuously calculated the angle of the door with a moving average over the last 5 samples. 
        Results are shown in Figure \ref{fig:real-microwave}. 
        We observe that the human demonstration is better than the one produced by the motion planner, which we credit to the fact that the motion of the door was heavily influenced by the dynamics of the revolute joint which the motion planner did not account for. Nonetheless, both policies converge to a similar final performance, with our method converging slightly faster. 
        Note the importance of the policy search phase: the motion planner alone is insufficient for performing the task efficiently.

        %\stnote{So one thing I was thinking was that since you are proposing this motion planning method in this paper, and you are also providing the human demonstration, you are incentivices to ``throw'' the human demonstration, either intentionally or subconsiously giving a suboptimal demonstration. Not saying you did this but might be worth adding a sentence or two saying that it was the authors (so acknowleding the limitation) and somehow objectively quantifying the quality of the demonstration (the human demonstration actually did complete the task or so)}
        
        Our second real-world task was to hit a ball off a tee as far as possible (Figure \ref{fig:real-tball}). Similar to our simulated task, the ball started at rest on top of the tee. The pose of the ball was used in the EstimateGrasp method. The object state was defined as the object's $y$ position relative to its initial pose. 
        %This experiment used DMP policies initialized with LWR and optimized with PI$^2$-CMA. 
        We placed scotchlite-reflective tape on the surface of the ball and conducted our experiments within an OptiTrack motion-capture cage to track the object pose. We observe that when using a motion planner to hit the ball, it moves the bat in a linear motion to make contact, therefore transferring only horizontal motion to the ball. We qualitatively observe that during policy search, the robot learns a dynamic policy that accounts for the dynamics of the ball by applying force under the ball to ``scoop'' the ball upwards and forwards.

\section{RELATED WORK}To our knowledge, our method is the first to %\ernote{\sout{use kinematic models of the robot and objects to initialize motor skill learning.}}
use an object’s estimated kinematics in conjunction with a known robot dynamics model to bootstrap motor policy learning, and we discover and discuss important problems that are only introduced when leveraging policy-learning algorithms, behavioral-cloning, and motion planning algorithms to do so. 
In this section, we discuss relevant approaches to motor skill learning.

Recently, Model-Predictive Control (MPC) has been used in the context of imitation learning and reinforcement learning to address the high sample complexity of policy search \cite{kahn2017plato, pan2017agile}. These approaches require a priori object dynamics, or human demonstrations to fit learned models; in constrast, our approach requires only object kinematics, which are much more readily estimated from visual data at runtime \cite{benny, li2019category}.  As such, our approaches enables the learning of manipulation skills to be more autonomous than existing MPC-based methods. 
    \citet{tosun2019pixels} proposed a neural network model for generating trajectories from images, using a motion planner during training to enable the robot to generate a trajectory with a single forward pass at runtime. While this approach uses a motion planner for behavior cloning, it stops short of optimization to improve the resulting policy. In constrast, our method uses object kinematics to produce initial trajectories, while \citet{tosun2019pixels} only use the robot's kinematic model, which is insufficient when the task is to manipulate an object to a specific joint configuration.  
    
    While classic robot motor learning papers \cite{atkeson1997robot} leverage the known kinodynamics of the robot, they do not discuss kinematics of external objects or grasp candidates to bootstrap motor policies for object manipulation. We emphasize that we cannot form dynamic plans in the problem setting we are interested in: objects with unknown a priori dynamics.
    
    \citet{kurenkov2019ac} proposed training an initially random RL policy with an ensemble of task-specific, hand-designed heuristics. This  improves learning but the initial policy is still random, yielding potentially unsafe behavior on real hardware, and delaying convergence to a satisfying policy. By contrast, we choose to initialize the policy with demonstrations from a kinematic planner, ensuring feasibility, safety, and rapid learning. Moreover, we argue that motion planning is the principled heuristic to use to accelerate learning, as it is capable of expressing manually programmed heuristics like reaching and pulling. Finally, our approach can use the existing estimated object kinematics to provide a principled reward signal for model-free reinforcement learning.
    
    Recently, residual reinforcement learning approaches have been developed which learn a policy superimposed on hand-designed or model-predictive controllers \cite{silver2018residual, johannink2019residual}. Our method is compatible with these approaches, where demonstrations from the motion planner can be used as a base policy on top of which a residual policy can be learned based on kinematic rewards. These methods typically suffer from the same limitations as MPC-based methods mentioned above. %\banote{TODO: emphasize these approaches would need object dynamics or demonstrations up front to solve our problem}

    %\banote{TODO: differentiate ourselves. \\
    %emphasize GPS still needs human demonstrations at the start to initialize models; GPS is used during policy optimization; our method is only used for exploration at the start of learning, not as a guide during optimization. We could literally use GPS as the optimization algorithm and still be novel.}

    % \citet{levine2013guided} presented Guided Policy Search (GPS), which uses trajectory optimization to guide policy search for model-free learning into high-reward regions via importance sampling. Our proposed framework is compatible with GPS's usage of importance sampling for guiding direct policy search to reduce sample complexity. However, our work differs in that we use kinematic models and sample-based motion planners instead of differentiable dynamic programming (DDP) to generate guiding samples, and we use a reward function parameterized by the kinematic model states instead of robot states. Our approach is particularly attractive in contrast to \citet{levine2013guided} because we can directly leverage existing state-of-the-art work in object kinematic estimation \cite{benny, li2019category} to unify demonstration generation and model-free reinforcement learning.
    
    Guided Policy Search (GPS) \cite{levine2013guided} uses LQR to guide policy search into high-reward regions of the state-space. The models employed are fundamentally local approximations, and thus would benefit greatly from a wealth of suboptimal demonstrations from the outset (as made evident by \citet{chebotar2017path}). GPS is one of the state-of-the-art algorithms we expect to be used within our framework as the policy search implementation (Section \ref{sec:ps}). A critical distinction between our work and GPS is the notion of planning trajectories in object configuration spaces and reasoning about grasp candidates to achieve a desired manipulation. This is done using information available apriori, and thus is immediately capable of generating high-value policies, whereas GPS is estimating dynamics models given observed data (obtained either from demonstration or random initialization). In the absence of a human demonstrator, our method would provide far more useful data at the outset of learning than running a naively initialized linear-gaussian controller (as evidenced by our comparisons to random initialization). The ideas proposed in our paper are distinct from those put forth in GPS: we present a method for obtaining demonstrations under certain conditions in the absence of a human.

    Most similar to our line of work are those that use sample-based motion planners for improved policy learning. \citet{jurgenson2019harnessing} harness the power of reinforcement learning for neural motion planners by proposing an augmentation of Deep Deterministic Policy Gradient (DDPG) \cite{lillicrap2015continuous} that uses the known robot dynamics to leverage sampling methods like RRT* to reduce variance in the actor update and provide off-policy exploratory behavior for the replay buffer. However, \citet{jurgenson2019harnessing} are only able to address domains where they can assume good estimates of the dynamics model, such as producing free-space motions to avoid obstacles. Our setting, in contrast, focuses on object manipulation, where dynamics are not readily available, but are critical for learning good policies.  \citet{jiang2019task} address learning to improve plans produced by a motion planner, but do not bootstrap closed-loop policies. Motion planners aren't expressive enough to leverage the dynamics in object-manipulation tasks, especially in the presence of unknown dynamics, and traditionally are unable to handle perceptual data like RGB images. Our method, on the other hand, enables motion planning to bootstrap policies that are more expressive than the original planner.

\section{CONCLUSION}
\label{sec:conclusion}
	We have presented a method that uses kinematic motion planning to bootstrap robot motor policies. By assuming access to a potentially noisy description of the object kinematics, we are able to autonomously generate initial demonstrations that perform as well as human demonstrations, but do not require a human, resulting in a practical method for autonomous motor skill learning.   
	
	Our methodology is agnostic to the motion planner, motor policy class, and policy search algorithm, making it a widely applicable paradigm for learning robot motor policies. We demonstrate the power of our methodology by bootstrapping different policy classes with demonstrations from humans and a motion planner, and learn motor policies for three dynamic manipulation tasks: closing a microwave door, opening a drawer, and hitting a ball off a tee.  Our framework is the first to enable robots to autonomously bootstrap and improve motor policies with model-free reinforcement learning using only a partially-known kinematic model of the environment. 
	
\section*{Acknowledgement}
This research was supported by NSF CAREER Award 1844960 to Konidaris, and by the ONR under the PERISCOPE MURI Contract N00014-17-1-2699. 
Disclosure: George Konidaris is the Chief Roboticist of Realtime Robotics, a robotics company that produces a specialized
motion planning processor.

\bibliographystyle{plainnat}
\bibliography{bibliography}

\end{document}